\documentclass[11pt]{article}
\usepackage[preprint]{acl}
\usepackage{times}
\usepackage{latexsym}
\usepackage[T1]{fontenc}
\usepackage{amsmath}
\usepackage{amssymb}
\usepackage{booktabs}
\usepackage[utf8]{inputenc}
\usepackage{microtype}
\usepackage{inconsolata}
\usepackage{graphicx}
\usepackage{algorithm}
\usepackage{algpseudocode}
\usepackage{multirow}
\usepackage{tikz}
\usetikzlibrary{arrows.meta,positioning,fit,calc}
\usepackage{fancyvrb}
\usepackage{placeins}
\usepackage{fvextra}
\usepackage{float}
\usepackage[most]{tcolorbox}
\tcbuselibrary{breakable,listings}
\usepackage{upquote}
\usepackage{needspace}

\newtcblisting{PromptTemplate}[2][]{
  enhanced,
  breakable,
  listing only,
  colback=black!2,
  colframe=black!35,
  colbacktitle=black!8,
  coltitle=black,
  fonttitle=\bfseries\small,
  title={#2},
  boxrule=0.45pt,
  arc=2pt,
  left=1.2mm,
  right=1.2mm,
  top=1mm,
  bottom=1mm,
  before skip=6pt,
  after skip=8pt,
  listing options={
    basicstyle=\ttfamily\scriptsize,
    breaklines=true,
    breakatwhitespace=false,
    columns=fullflexible,
    keepspaces=true,
    showstringspaces=false,
    upquote=true
  },
  #1
}
\DefineVerbatimEnvironment{PromptBlock}{Verbatim}{
  breaklines=true,
  breakanywhere=true,
  fontsize=\scriptsize,
  baselinestretch=0.92,
  xleftmargin=0pt,
  xrightmargin=0pt
}

\title{RippleMem: From Isolated Retrieval to Associative Recollection for Long-Term Agent Memory}

\author{
Jingbo Ji\textsuperscript{1} \quad
Lingyi Li\textsuperscript{2} \quad
Xilong Cheng\textsuperscript{1} \quad
Yuhao Zhou\textsuperscript{1} \\
\bfseries Wenji Zhang\textsuperscript{1} \quad
\bfseries Yuting Tan\textsuperscript{1} \quad
\bfseries Yunxiao Qin\textsuperscript{1,3,} \thanks{Corresponding author} \\
\textsuperscript{1}Communication University of China, Beijing, China. \\
\textsuperscript{2} Zhilian Yinghe Technology Co., Ltd., Beijing, China\\
\textsuperscript{3}State Key Laboratory of Media Convergence and Communication, Beijing, China \\
\texttt{cocomilk23@mails.cuc.edu.cn, qinyunxiao@cuc.edu.cn}
}
\hypersetup{
  pdftitle={RippleMem: From Isolated Retrieval to Associative Recollection for Long-Term Agent Memory},
  pdfauthor={Jingbo Ji, Lingyi Li, Xilong Cheng, Yuhao Zhou, Wenji Zhang, Yuting Tan, Yunxiao Qin}
}

\begin{document}
\maketitle

\begin{abstract}
LLM-based agents increasingly rely on external memory to support long-horizon reasoning and interaction. 
However, the main bottleneck is not simply storing past experience, but recovering the right set of evidence when relevant information is distributed across many interactions. 
Existing approaches struggle with this access problem. 
Full-context methods require noisy long-context search, flat retrieval often returns isolated and incomplete records, and graph-based memory systems can be expensive to construct while compressing rich event context. 
We introduce \textbf{RippleMem}, a long-term memory system that replaces one-shot retrieval with adaptive associative recollection. 
Inspired by cue-dependent episodic retrieval and associative completion, RippleMem stores interaction history as cue-rich episodic memory units and organizes them in an event-centric memory graph. 
Given a query, it first recalls relevant memory anchors through hybrid cues, then expands from these anchors along semantic and structural associations to recover missing supporting evidence. In this way, initially recalled memories serve not only as answer context, but also as cues for completing the evidence needed to answer. Experiments on \textbf{LoCoMo} and \textbf{LongMemEval-S} show that RippleMem achieves the best overall performance across evaluated settings, improving LLM-as-a-Judge accuracy by \textbf{3.95\%} on LoCoMo and up to \textbf{11.87\%} on LongMemEval-S, while reducing graph construction cost by about \textbf{30$\times$}.
\end{abstract}

\section{Introduction}

Large language models (LLMs) are increasingly deployed as agents that must reason, plan, and interact over long horizons~\citep{park2023generative,NEURIPS2023_1b44b878,yao2023react}. 
In such settings, memory becomes an increasingly important capability rather than merely an auxiliary component, since an agent often needs to preserve useful information from past interactions and recover it as evidence when later queries depend on earlier events, preferences, or decisions~\citep{Hatalis_Christou_Myers_Jones_Lambert_Amos-Binks_Dannenhauer_Dannenhauer_2024,zhang2025survey}. 
In long-horizon interaction, however, the information needed to answer a query is often not contained in a single past utterance or memory record.
It may be distributed across temporally distant sessions, mixed with routine dialogue, and become useful only when multiple fragments are recovered together.
% Thus, a central challenge for long-term memory is not merely storing past experience, but recovering an answerable set of evidence from distributed traces.
Thus, a central challenge for long-term memory is not merely storing past experience, but recollecting an answerable set of evidence from distributed traces.

Recent long-term memory systems have therefore explored ways to make stored interaction histories more accessible at query time.
Some systems store past interactions as compact memory records and retrieve them with semantic or query-aware mechanisms, supporting efficient personalization over long histories~\citep{Zhong_Guo_Gao_Ye_Wang_2024,chhikara2025mem0buildingproductionreadyai,liu2026simplememefficientlifelongmemory}.
Memory operating systems and lifecycle-based frameworks further maintain, update, and reorganize memories over time, providing a more stable substrate for long-horizon access~\citep{li2025memosmemoryosai,hu2026evermemosselforganizingmemoryoperating}.
A complementary line introduces explicit structure or association, using graphs, temporal links, or recollection-style retrieval to let memory access move beyond independent record lookup~\citep{xu2025amemagenticmemoryllm,rasmussen2025zeptemporalknowledgegraph,zhang2026evokingusermemorypersonalizing}.

Despite these advances, memory access can still fail to assemble an answerable evidence set.
Flat query-centered lookup may stop at the most directly matched record, while graph expansion without an explicit evidence need may traverse nearby but non-supporting memories. 
In such cases, the system can miss supporting evidence that has been stored but is not recovered together with the initially surfaced memory.

\begin{figure}[t]
    \centering
    \includegraphics[width=\columnwidth]{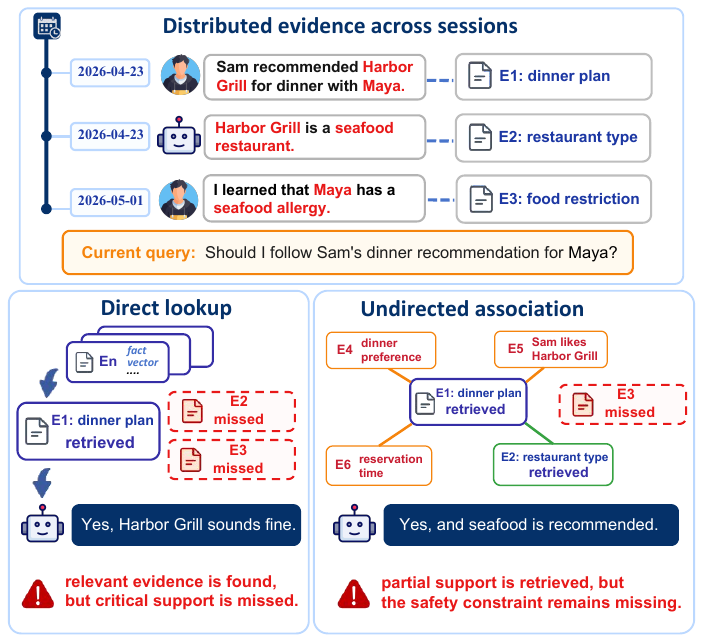}
    \vspace{-22pt}
   \caption{Failure modes in long-term memory access. Answer-critical evidence may be stored across sessions, while direct lookup or undirected association can still fail to assemble the missing support needed for an appropriate answer.}
    \label{fig:motivation}
\end{figure}

Such failures raise a central question: when an initially retrieved memory is relevant but incomplete, how should memory access continue to recover the missing support needed for answering? As illustrated in Figure~\ref{fig:motivation}, the query requires not only a dinner-plan memory, but also separately stored evidence about the guest's food restriction and the restaurant type. 
Direct lookup may stop too early, while undirected association may surface related traces without ensuring that the missing evidence role is filled. Motivated by this observation, we argue that long-term memory access should be evidence-conditioned: recalled memories should serve as cues for recovering missing support, rather than as the endpoint of retrieval.

To operationalize this idea, we propose \textbf{RippleMem}, a long-term memory system that treats memory access as adaptive associative recollection. 
The name reflects the central mechanism: like ripples spreading from an initial point of contact, RippleMem starts from initially recalled memories and expands locally through linked episodic memories to recover missing evidence. 
This design is informed by cognitive accounts that organize experience into event-like units~\citep{zacks2007event,baldassano2017discovering} and view recall as cue-dependent and associative~\citep{tulving1973encoding,norman2003modeling}. We use these accounts as design intuitions rather than mechanistic claims. RippleMem operationalizes them through a write--read design: the write phase stores interaction history as cue-rich episodic memory units linked by semantic and structural associations, while the read phase uses initially recalled memories as cues for recovering additional support. This moves memory access beyond one-shot query matching and unguided graph traversal toward evidence completion over stored event memories.

In summary, our contributions are as follows:
\begin{itemize}
    \item We formulate long-term memory access as an evidence recovery problem: relevant history may be stored but still fail to be recovered as an answerable evidence set when support is distributed across interactions. This framing highlights a limitation of one-shot query--memory matching.
    \item We propose \textbf{RippleMem}, an event-centric long-term memory system that combines cue-rich episodic memory construction with anchor-local associative recollection. Its key design lets recalled memories serve as both answer context and cues for recovering missing support through semantic and structural associations.
    \item We show that RippleMem achieves the best overall performance on LoCoMo and LongMemEval-S, relatively improving LLM-as-a-Judge accuracy by \textbf{3.95\%} on LoCoMo and up to \textbf{11.87\%} on LongMemEval-S over the strongest baselines, while reducing graph construction cost by about \textbf{30$\times$} compared with graph-based memory baselines.
\end{itemize}

\section{Related Work}

\subsection{Memory Mechanisms for LLM Agents}

\paragraph{Long-context and retrieval-augmented memory.}
Large language models are constrained by finite context windows, and recent work improves long-context processing through extended context modeling and long-context prompting~\citep{dai-etal-2019-transformer,bai-etal-2024-longbench,beltagy2020longformerlongdocumenttransformer}. However, longer context does not guarantee reliable memory access, as models remain vulnerable to evidence degradation and lost-in-the-middle effects~\citep{liu-etal-2024-lost}. Retrieval-augmented generation externalizes memory through document or chunk retrieval~\citep{lewis2021retrievalaugmentedgenerationknowledgeintensivenlp,gao2024retrievalaugmentedgenerationlargelanguage,fan2024survey}, but its effectiveness depends on whether the retriever surfaces evidence that may be distributed across turns, sessions, and time~\citep{yu-etal-2024-chain,sorodoc-etal-2025-garage}. Graph-based RAG further organizes external knowledge for relational
access, including n-ary hypergraph representation in HyperGraphRAG,
relational-path pruning in PathRAG, and graph foundation model-based reasoning in G-reasoner~\citep{edge2025localglobalgraphrag,
luo2025hypergraphrag,
chen2026pathrag,
luo2026greasoner}. This line primarily targets document- or knowledge-centric graphs, while RippleMem studies evidence recovery over evolving episodic
interaction memory.

\paragraph{Long-term memory systems.}
A substantial body of work builds long-term memory for LLM agents by transforming interaction history into searchable memory entries~\citep{lee2024humaninspiredreadingagentgist,tan-etal-2025-prospect}. MemGPT~\citep{packer2024memgptllmsoperatingsystems} treats memory as a virtual context space and manages information through memory paging. MemoryBank~\citep{Zhong_Guo_Gao_Ye_Wang_2024} stores user-specific information across interactions, while Mem0~\citep{chhikara2025mem0buildingproductionreadyai} emphasizes concise long-term memory extraction for downstream retrieval. SimpleMem~\citep{liu2026simplememefficientlifelongmemory} further improves memory quality through semantic compression, structured indexing, and query-aware retrieval planning. Recent systems also organize memory with explicit structure.
MemTree~\citep{ICLR2025_0382cb76} supports coarse-to-fine access
through hierarchical memory organization.
Memory operating systems such as MemOS and
EverMemOS~\citep{li2025memosmemoryosai,
hu2026evermemosselforganizingmemoryoperating} manage memory as a system-level resource. Graph-based architectures such as A-MEM~\citep{xu2025amemagenticmemoryllm} and Zep~\citep{rasmussen2025zeptemporalknowledgegraph} link memories through structured associations or temporal knowledge-graph relations. While these approaches provide practical substrates for persistent memory, they leave open how memory access should continue when initially retrieved evidence is relevant but incomplete.

\subsection{Episodic and Associative Memory Access}

\paragraph{Cognitive accounts of episodic recollection.}
Human episodic memory offers a lens for long-term agent memory. Episodic memory concerns event-specific experiences and their contextual details~\citep{tulving1972episodic}. Event segmentation accounts further suggest that continuous experience is organized into discrete event-like units, which support later understanding and recall~\citep{zacks2007event,baldassano2017discovering}. Recollection also depends on binding item information with contextual features such as time, place, and surrounding context~\citep{davachi2006item,yonelinas2019contextual}. The encoding specificity principle suggests that retrieval succeeds when current cues overlap with information encoded during the original experience~\citep{tulving1973encoding}. Beyond direct cue matching, hippocampal indexing and pattern-completion accounts suggest that partial cues can reactivate associated traces and recover additional information~\citep{norman2003modeling}. RippleMem does not model human memory mechanistically, but adopts these ideas as design intuitions for event-level memory construction, cue-rich contextual binding, and associative recollection.

\paragraph{Associative access in memory systems.}
Recent memory systems explore structured access beyond one-shot matching. M-Flow~\citep{mflow2026github} organizes memories into a cone-shaped multi-granularity graph and scores episode bundles through graph-routed path propagation. MemGAS~\citep{xu2025singlemultigranularitylongtermmemory} constructs multi-granularity memory representations and selects among them for context construction. REMem~\citep{shu2026remem} organizes
time-aware gists and time-scoped facts in a hybrid graph and uses an
agentic retriever for iterative evidence gathering. RF-Mem~\citep{zhang2026evokingusermemorypersonalizing} proposes a
retrieval-side familiarity--recollection mechanism that adapts between
direct retrieval and recollection-style expansion. These works show
that long-term memory access benefits from structure, association, and
adaptive retrieval. RippleMem is complementary to this line of work and focuses on how evidence already recovered during retrieval can serve as a cue for finding additional supporting memories.

\section{Method}

RippleMem is a long-term memory framework for recovering answer-supporting evidence from interaction history. It first constructs an event-centric memory substrate, and then uses this substrate to perform adaptive associative recollection at inference time.

% RippleMem is motivated by cue-dependent associative recollection. 
% It stores event memories with contextual cues, reflecting the role of item--context binding in episodic memory~\citep{tulving1972episodic,davachi2006item,yonelinas2019contextual}. 
% At read time, recalled memories act as anchors for recovering related support, consistent with partial-cue accounts of associative recollection~\citep{norman2003modeling}.
% This is a computational design principle rather than a mechanistic model of human memory.

\subsection{Overview}
\label{sec:method-overview}

\begin{figure*}[t]
    \centering
    \includegraphics[width=\textwidth]{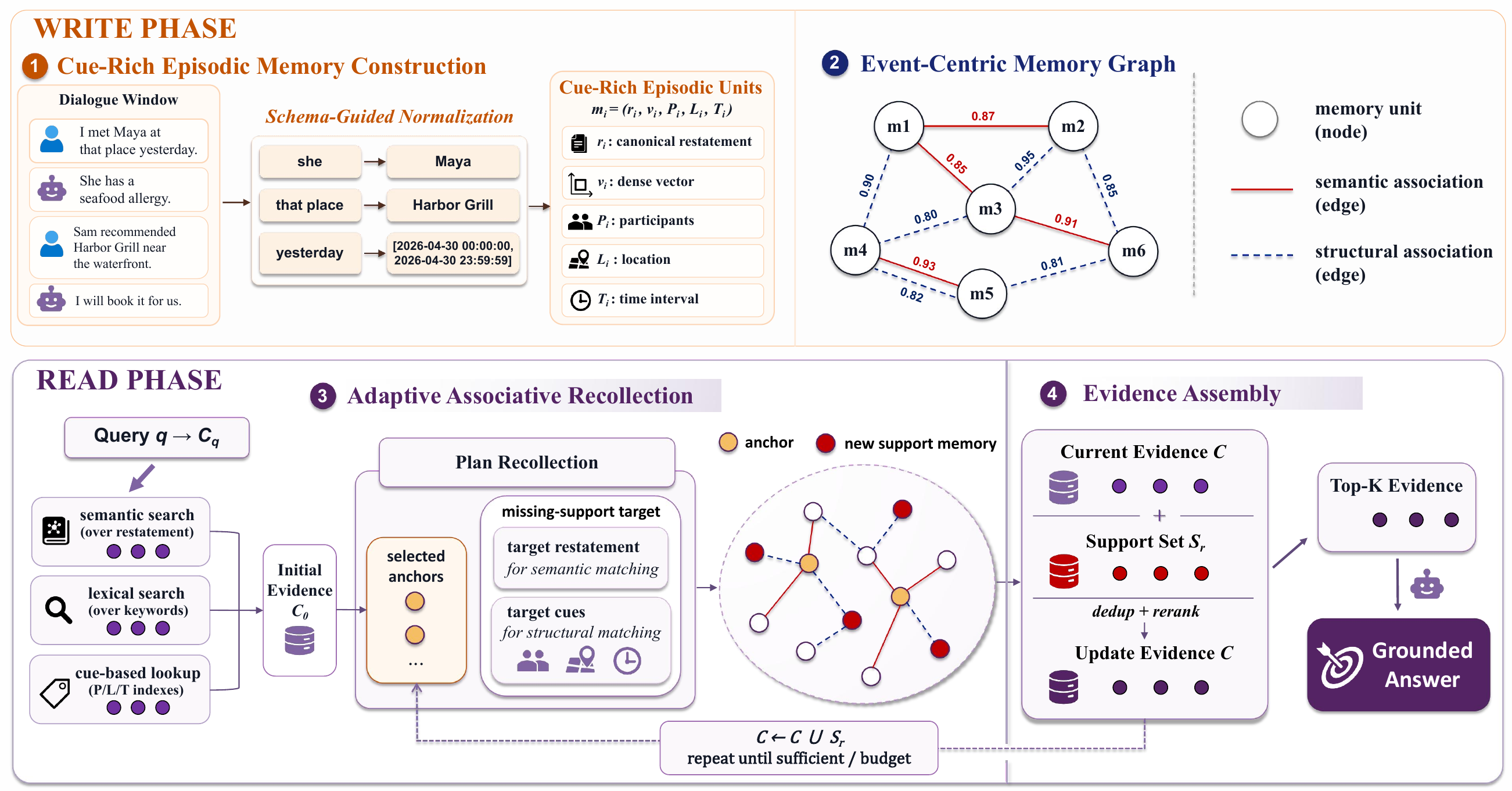}
    \vspace{-22pt}
     \caption{
    Illustration of the RippleMem framework, including cue-rich episodic memory construction, event-centric memory graph construction, adaptive associative recollection, and evidence assembly.
    In recollection, anchors define where to expand, while the missing-support target guides what support to recover.
    }
    \label{fig:ripplemem-framework}
\end{figure*}

Figure~\ref{fig:ripplemem-framework} illustrates the overall architecture of RippleMem, organized around a write--read design.

\textbf{Write Phase.}
The write phase builds the memory substrate for later recollection. \textbf{Cue-Rich Episodic Memory Construction} converts dialogue history into self-contained event memories with semantic representations and episodic cues, while the \textbf{Event-Centric Memory Graph} links these memories through semantic and structural associations. Together, they transform raw interaction history into a connected event memory space.

\textbf{Read Phase.}
The read phase uses this event memory space to recover evidence for a query. \textbf{Adaptive Associative Recollection} first recalls candidate evidence and then, when needed, follows local associations from salient recalled memories to recover missing support. \textbf{Evidence Assembly} consolidates the resulting memories into a compact evidence context for grounded response generation.

Overall, RippleMem treats memory access as controlled evidence completion rather than one-shot retrieval, with recalled memories serving as both candidate evidence and cues for recovering additional support.

\subsection{Cue-Rich Episodic Memory Construction}
\label{sec:memory-construction}

As illustrated in the write phase of Figure~\ref{fig:ripplemem-framework}, RippleMem is motivated by event segmentation accounts~\citep{zacks2007event,baldassano2017discovering} and converts interaction history into cue-rich episodic memory units. Given a dialogue trace, we process it as contiguous turn-level windows $\{W_t\}$ with local overlap to preserve continuity across window boundaries. Each window is passed to a schema-guided LLM extractor, which returns a JSON list of memory units and may return an empty list when the window contains no durable event, preference, commitment, observation, or plan worth storing. The outputs are validated against the memory schema before being written to the memory bank.

Each memory unit follows a fixed schema:
\begin{equation}
    m_i = (r_i, \mathbf{v}_i, P_i, L_i, T_i),
\end{equation}
where $r_i$ is a canonicalized event restatement that can be interpreted independently of the original dialogue window, $\mathbf{v}_i$ is the dense representation of $r_i$, and $P_i$, $L_i$, and $T_i$ denote grounded participants, locations, and temporal cues or intervals when available. 

To retain the contextual bindings central to episodic recollection~\citep{davachi2006item,yonelinas2019contextual}, the structured cue fields preserve who, where, and when information for later cue-based access and association. The extractor is instructed to resolve speaker-dependent references into explicit participant names, anchor relative time expressions to absolute intervals only when supported by the dialogue timestamp, and split multi-event windows into independently retrievable memory units. If a participant, location, or time expression cannot be grounded, the corresponding field is left empty rather than inferred.

The dense vector $\mathbf{v}_i$ supports semantic access, while $(P_i,L_i,T_i)$ expose episodic cues that can later participate in structured and cross-cue association. Thus, the write phase does not merely compress dialogue into shorter summaries; it constructs cue-addressable event memories intended to support associative recollection.

\subsection{Event-Centric Memory Graph}
\label{sec:memory-graph}

After memory construction, RippleMem organizes the memory bank as a sparse weighted event graph $G=(\mathcal{M},\mathcal{E})$.
As shown in Figure~\ref{fig:ripplemem-framework}, each node is a memory unit, and $\mathcal{E}=\mathcal{E}_{\mathrm{sem}}\cup\mathcal{E}_{\mathrm{str}}$ contains typed links retained after association scoring. Each retained edge carries its corresponding association score as the edge weight and serves as a potential recollection path between two memory units.

\paragraph{Semantic association.}
Semantic association captures meaning-level proximity between memory restatements. Since each memory unit has a dense representation $\mathbf{v}_i$, RippleMem computes
\begin{equation}
    s_{\mathrm{sem}}(i,j)=\cos(\mathbf{v}_i,\mathbf{v}_j).
\end{equation}
This channel preserves conceptual continuity in the graph, enabling a recalled event to cue semantically related memories during recollection.

\paragraph{Structural association.}
Structural association captures shared grounded episodic cues. For a pair of memory units, RippleMem computes a cue-similarity vector
$\mathbf{u}_{ij}=[u^P_{ij},u^L_{ij},u^T_{ij}]$.
Here $u^P_{ij}$ and $u^L_{ij}$ are Jaccard overlaps over canonicalized participant and location fields. For temporal cues, RippleMem assigns higher compatibility to overlapping or nearby intervals:
\begin{equation}
u^T_{ij} =
\begin{cases}
1, & T_i \cap T_j \neq \emptyset,\\
\exp(-\Delta(T_i,T_j)/\tau), & \text{otherwise},
\end{cases}
\end{equation}
where $\Delta(T_i,T_j)$ is the gap between two non-overlapping intervals and $\tau$ is a fixed temporal decay scale.

RippleMem aggregates cue-level similarities into a structural compatibility score over comparable cue types:
\begin{equation}
s_{\mathrm{str}}(i,j)
=
\frac{
\sum_{x\in\{P,L,T\}} \beta_x \mathbb{I}_x u^x_{ij}
}{
\sum_{x\in\{P,L,T\}} \beta_x \mathbb{I}_x
},
\end{equation}
where $\mathbb{I}_x$ indicates whether cue type $x$ is present and comparable for both memory units, and $\beta_x$ denotes a fixed cue weight. Pairs with no comparable cue type are not assigned a structural score.

\paragraph{Sparse graph construction.}
Graph construction is incremental and sparse. When a new memory unit is inserted, RippleMem obtains a bounded candidate pool from semantic nearest-neighbor search and cue-based indexes over participants, locations, and time. It then scores candidates through the two association channels. Semantic links are retained from candidates whose $s_{\mathrm{sem}}$ exceeds a score threshold, up to $K_{\mathrm{sem}}$ links; structural links are retained analogously using $s_{\mathrm{str}}$, up to $K_{\mathrm{str}}$ links. A pair of memory units may be connected through either channel or both. During anchor-local recollection, the two typed channels can be traversed separately and then merged, allowing the system to recover memory units related by meaning, grounded episodic cues, or both.

\subsection{Adaptive Associative Recollection}
\label{sec:recollection}

\paragraph{Hybrid initial recall.}
As shown in the read phase of Figure~\ref{fig:ripplemem-framework}, RippleMem begins recollection by extracting retrieval cues $c_q$ from a query $q$, including semantic, lexical, and grounded episodic cues. 
Consistent with the encoding-specificity view of retrieval~\citep{tulving1973encoding}, these cues instantiate three complementary recall views:
\begin{equation}
    C_0 = C_{\mathrm{sem}} \cup C_{\mathrm{lex}} \cup C_{\mathrm{cue}},
\end{equation}
where $C_{\mathrm{sem}}$, $C_{\mathrm{lex}}$, and $C_{\mathrm{cue}}$ are obtained through semantic search, lexical matching, and structured cue matching, respectively. We initialize the current evidence state as $C \leftarrow C_0$ and update it after each recollection round.

\paragraph{Memory-anchor planning.}
Given the current evidence state $C$, RippleMem invokes a schema-constrained recollection controller $\Pi_{\mathrm{rec}}$:
\begin{equation}
    (d_r,A_r,g_r,s_r)=\Pi_{\mathrm{rec}}(q,c_q,C).
\end{equation}
The output specifies whether to continue ($d_r$), the selected anchors ($A_r\subseteq C$), the missing-support target ($g_r$), and an optional stop reason ($s_r$). Given the current evidence state $C$, the controller decides whether further recollection is needed. If so, it selects anchors that are central to the query or likely to evoke missing support through graph associations, and defines $g_r$ using a target restatement and optional structured cues.

\paragraph{Anchor-local expansion.}
Given a valid plan, RippleMem expands only from the selected anchors within a bounded graph neighborhood:
\begin{equation}
    U_r =
    \{m \in \mathcal{M}\setminus C
    \mid d_G(m,A_r) \le h\}.
\end{equation}
Here $d_G$ is computed on the union graph induced by semantic and structural edges, while subsequent selection retains channel-specific provenance. Expansion proceeds over the two graph channels. For each selected anchor, RippleMem collects candidates reachable within $h$ hops through semantic edges and structural edges, excluding memories already in $C$. Candidates are then scored against the missing-support target: the target restatement is used for semantic matching, and the target cues are used for structural matching. The top candidates from the two channels are merged by memory identity to form the round-level support set $S_r$.

The evidence state is updated as $C \leftarrow C \cup S_r$. When candidates have similar target-match scores, RippleMem prefers candidates that are closer to the anchor and connected through stronger weighted paths. This loop continues until the evidence is sufficient, no new support is recovered, or the round budget is reached, as summarized in Algorithm~\ref{alg:recollection}.

\begin{algorithm}[t]
\small
\caption{Adaptive associative recollection}
\label{alg:recollection}
\begin{algorithmic}[1]
\Require query $q$, query cues $c_q$, memory graph $G$
\Require initial evidence $C_0$, hop limit $h$, round budget $R$
\State $C \gets C_0$
\For{$r=1$ to $R$}
    \State $(d_r,A_r,g_r,s_r) \gets \Pi_{\mathrm{rec}}(q,c_q,C)$
    \If{$d_r=\textsc{stop}$ or $A_r=\emptyset$}
        \State \textbf{break}
    \EndIf
    \State $U_r \gets \mathrm{LocalNeighbors}(G,A_r,h)\setminus C$
    \State $S_r \gets \mathrm{SelectSupport}(U_r,g_r,A_r)$
    \If{$S_r=\emptyset$}
        \State \textbf{break}
    \EndIf
    \State $C \gets C \cup S_r$
\EndFor
\State \Return $C$
\end{algorithmic}
\end{algorithm}

\subsection{Evidence Assembly}
\label{sec:evidence-assembly}

\paragraph{Evidence consolidation.}
As shown in the evidence assembly component of Figure~\ref{fig:ripplemem-framework}, RippleMem converts the final evidence state $C$ into a budgeted evidence context for response generation. It first consolidates memories by memory-unit identity. If the same memory unit is reached through multiple recall views, anchors, or anchor-local expansion paths, RippleMem keeps a single copy and merges its provenance information, producing a consolidated memory set $\bar{C}$.

RippleMem then applies a deterministic source-aware ordering to $\bar{C}$. 
The score combines query--memory alignment, retrieval provenance, and anchor status:
\begin{equation}
    \rho(m)=
    \lambda_q a(q,m)
    + \lambda_p \pi(m)
    + \lambda_a \mathbb{I}_{\mathrm{anc}}(m),
\end{equation}
where $a(q,m)$ is the normalized semantic alignment between the query and the memory restatement, $\pi(m)$ summarizes retrieval-source rank, source score, and expansion-path support, and $\mathbb{I}_{\mathrm{anc}}(m)$ indicates whether $m$ served as a memory anchor. 
All weights are fixed before evaluation; the exact instantiation is provided in Appendix~\ref{app:implementation-settings}.

The final evidence context $E_K$ consists of the top-$K$ memory units under this ordering, where $K$ is a memory-count budget fixed before evaluation and kept unchanged across benchmarks. The answer generator receives the original query together with $E_K$ and produces the final response conditioned on the assembled memories.

\section{Experiments}
We evaluate RippleMem through main benchmark comparisons, ablation studies, and phase-level efficiency analysis. Detailed dataset statistics, implementation settings, and qualitative cases are provided in the Appendix.

\subsection{Experimental Setup}
\label{sec:experimental-setup}

\paragraph{Datasets.}
We evaluate RippleMem on LoCoMo~\citep{maharana-etal-2024-evaluating} and LongMemEval-S~\citep{wu2025longmemeval}, two long-term conversational memory benchmarks covering ultra-long dialogues and full-history user-agent interactions~\citep{he-etal-2025-madial,tan-etal-2025-membench,wang-zhao-2024-tram,chu-etal-2024-timebench}. Detailed dataset statistics and question categories are provided in Appendix~\ref{app:evaluation-settings}. We additionally evaluate RippleMem on EverMemBench~\citep{hu2026evermembench}, a dynamic multi-party and multi-group conversational memory benchmark, with the full protocol and results reported in Appendix~\ref{app:evermembench}.

\begin{table*}[!t]
\centering
\footnotesize
\setlength{\tabcolsep}{4.0pt}
\renewcommand{\arraystretch}{1.12}
\resizebox{\textwidth}{!}{
\begin{tabular}{l|ccc|ccc|ccc|ccc|ccc}
\toprule
\multirow{2}{*}{Method}
& \multicolumn{3}{c|}{Multi-Hop}
& \multicolumn{3}{c|}{Temporal}
& \multicolumn{3}{c|}{Open Domain}
& \multicolumn{3}{c|}{Single Hop}
& \multicolumn{3}{c}{Average} \\
& F1 $\uparrow$ & B1 $\uparrow$ & J $\uparrow$
& F1 $\uparrow$ & B1 $\uparrow$ & J $\uparrow$
& F1 $\uparrow$ & B1 $\uparrow$ & J $\uparrow$
& F1 $\uparrow$ & B1 $\uparrow$ & J $\uparrow$
& F1 $\uparrow$ & B1 $\uparrow$ & J $\uparrow$ \\
\midrule
Full-Context
& 32.06 & 23.63 & 61.70
& 44.39 & 33.23 & 50.78
& 21.60 & 17.88 & 53.13
& 50.72 & 44.88 & 81.57
& 44.17 & 36.83 & 69.74 \\
Mem0
& 32.42 & 24.70 & 62.06
& 47.91 & 42.03 & 64.49
& 20.20 & 15.56 & 53.13
& 39.31 & 34.70 & 62.54
& 38.65 & 33.20 & 62.27 \\
Mem0$^\mathrm{g}$
& 34.43 & 25.91 & 68.44
& 51.41 & 44.47 & 64.80
& 22.68 & 16.93 & 57.29
& 41.84 & 36.81 & 66.59
& 41.28 & 35.17 & 65.97 \\
Zep
& 30.64 & 22.89 & 66.31
& 49.05 & 37.97 & 70.72
& 22.55 & 18.60 & 60.42
& 49.08 & 42.87 & 84.30
& 43.98 & 36.68 & 76.69 \\
MemGAS
& 16.70 & 13.09 & 64.89
& 15.63 & 10.01 & 48.91
& 12.15 & 9.04 & 56.25
& 23.10 & 13.92 & 71.58
& 19.69 & 12.65 & 64.68 \\
M-Flow
& 33.76 & 26.39 & 72.34
& 42.83 & 36.31 & 61.68
& 15.51 & 12.10 & 61.46
& 40.48 & 31.69 & 83.74
& 38.18 & 30.46 & 75.67 \\
REMem
& 26.28 & 20.09 & 71.99
& 35.77 & 24.63 & 82.55
& 23.69 & 19.60 & 61.46
& 37.54 & 28.45 & 83.47
& 34.24 & 25.57 & 79.81 \\
SimpleMem
& 37.12 & 29.98 & \textbf{78.01}
& 56.92 & 42.77 & 76.01
& 24.75 & 19.77 & 63.54
& 55.45 & 49.21 & 89.42
& 50.48 & 42.51 & 82.92 \\
RF-Mem$^\dagger$
& 37.04 & 29.00 & 75.89
& 59.62 & 45.15 & 80.69
& \textbf{26.54} & \textbf{21.87} & 68.75
& 54.14 & 48.00 & 89.42
& 50.43 & 42.30 & 83.83 \\
RippleMem
& \textbf{38.58} & \textbf{31.12} & 77.67
& \textbf{62.37} & \textbf{47.29} & \textbf{85.67}
& 25.71 & 20.93 & \textbf{70.83}
& \textbf{56.44} & \textbf{49.79} & \textbf{92.75}
& \textbf{52.49} & \textbf{44.05} & \textbf{87.14} \\
\bottomrule
\end{tabular}
}
\vspace{-6pt}
\caption{Main LoCoMo results. We report F1, BLEU-1 (B1), and LLM-as-a-Judge accuracy (J). RF-Mem$^\dagger$ uses RippleMem's extracted memory units with a matched per-question evidence budget. Bold denotes the best result.}
\label{tab:locomo-main}
\end{table*}

\paragraph{Baselines.}
We compare RippleMem with representative long-context and memory-augmented baselines. On LoCoMo, we include Full-Context, Mem0 and Mem0$^\mathrm{g}$~\citep{chhikara2025mem0buildingproductionreadyai}, Zep~\citep{rasmussen2025zeptemporalknowledgegraph}, MemGAS~\citep{xu2025singlemultigranularitylongtermmemory}, M-Flow~\citep{mflow2026github}, REMem~\citep{shu2026remem}, SimpleMem~\citep{liu2026simplememefficientlifelongmemory}, and RF-Mem~\citep{zhang2026evokingusermemorypersonalizing}. On LongMemEval-S, we additionally compare with reported results from LightMem~\citep{fang2026lightmemlightweightefficientmemoryaugmented}, MemU~\citep{memu2025github}, MemOS~\citep{li2025memosmemoryosai}, and EverMemOS~\citep{hu2026evermemosselforganizingmemoryoperating}. Baseline hyperparameters follow the original papers, public implementations, or official recommendations when available; brief descriptions are provided in Appendix~\ref{app:baseline-descriptions}.

\paragraph{Metrics.}
For LoCoMo, we report F1, BLEU-1, and LLM-as-a-Judge accuracy. F1 and BLEU-1 measure lexical overlap with reference answers, while the judge score evaluates semantic correctness beyond exact surface matching. For LongMemEval-S, we report binary LLM-as-a-Judge accuracy following its accuracy-style evaluation protocol. Appendix~\ref{app:metric-aggregation} and Appendix~\ref{app:judge-protocol} provide metric aggregation and judge details.

\paragraph{Implementation Details.}
Unless otherwise specified, GPT-4.1-mini is the backbone LLM for memory extraction, query analysis, recollection planning, and answer generation, with all decoding temperatures set to zero. 
For LoCoMo, GPT-4.1-mini also serves as the judge, and Qwen3-Embedding-0.6B is used as the dense encoder. 
For LongMemEval-S, we follow two prior-aligned settings: SimpleMem-aligned with GPT-4.1-mini judge and Qwen3-Embedding-0.6B encoder, and EverMemOS-aligned with GPT-4o-mini judge and Qwen3-Embedding-4B encoder, taking baseline results from the corresponding prior work.

RippleMem uses fixed benchmark-specific budgets for memory construction, first-hop recall, graph expansion, and evidence assembly, as reported in Appendix~\ref{app:implementation-settings}. 
For RF-Mem, we use the memory units extracted by RippleMem and match RippleMem's per-question evidence budget to isolate the memory-access strategy.

\begin{table*}[!t]
\centering
\small
\setlength{\tabcolsep}{4.6pt}
\renewcommand{\arraystretch}{1.12}
% \resizebox{\textwidth}{!}{
\begin{tabular}{c l|ccccccc}
\toprule[1.2pt]
Setting & Method
& SS-User
& Multi-S
& SS-Pref
& Temp. Reas
& Know. Upd
& SS-Asst
& Overall \\
\midrule

\multirow{5}{*}{\shortstack[c]{\textit{SimpleMem}\\\textit{setting}}}
& Full-Context
& 47.14 & 30.08 & 60.00 & 27.06 & 41.03 & 32.14 & 35.40 \\
& Mem0
& 87.14 & 50.37 & 63.33 & 40.60 & 69.23 & 48.21 & 58.40 \\
& LightMem
& 88.57 & 47.37 & 76.67 & \textbf{85.71} & \textbf{92.30} & 21.43 & 69.20 \\
& SimpleMem
& 85.71 & 60.92 & 76.67 & 83.46 & 79.48 & 75.00 & 75.80 \\
& RippleMem
& \textbf{97.14} & \textbf{78.20} & \textbf{96.67} & 76.70 & 91.03 & \textbf{89.29} & \textbf{84.80} \\

\midrule

\multirow{6}{*}{\shortstack[c]{\textit{EverMemOS}\\\textit{setting}}}
& MemU
& 67.14 & 42.10 & 77.67 & 17.29 & 41.02 & 19.64 & 38.40 \\
& Zep
& 92.90 & 47.40 & 53.30 & 54.10 & 74.40 & 75.00 & 63.80 \\
& Mem0
& 82.86 & 63.15 & 90.00 & 72.18 & 66.67 & 26.78 & 66.40 \\
& MemOS
& 95.71 & 70.67 & \textbf{96.67} & 77.44 & 74.26 & 67.86 & 77.80 \\
& EverMemOS
& \textbf{97.14} & 73.68 & 93.33 & 77.44 & \textbf{89.74} & 85.71 & 83.00 \\
& RippleMem
& 95.71 & \textbf{80.45} & 83.33 & \textbf{84.21} & 88.46 & \textbf{94.64} & \textbf{86.60} \\

\bottomrule[1.2pt]
\end{tabular}
% }
\caption{
Main results on LongMemEval-S.
We report LLM-as-a-Judge accuracy (\%).
SS denotes single-session; Asst and Pref denote assistant and preference; Multi-S denotes multi-session; Know. Upd and Temp. Reas denote knowledge-update and temporal-reasoning questions.
Bold denotes the best result within each comparison group.
}
\label{tab:lme-main}
\end{table*}

\subsection{Main Results}
\label{sec:main-results}

Tables~\ref{tab:locomo-main} and~\ref{tab:lme-main} summarize the main results for LoCoMo and LongMemEval-S. We highlight three key observations.

\paragraph{Consistent overall gains across benchmarks.}
RippleMem achieves the best overall performance on both benchmarks. On LoCoMo, it obtains 52.49\% F1, 44.05\% BLEU-1, and 87.14\% LLM-as-a-Judge accuracy. Against the strongest baseline for each metric, RippleMem relatively improves F1 and BLEU-1 over SimpleMem by 3.98\% and 3.62\%, respectively, and judge accuracy over RF-Mem by 3.95\%. On LongMemEval-S, RippleMem achieves the best overall accuracy in both comparison groups, reaching 84.80\% under the SimpleMem evaluation setting and 86.60\% under the EverMemOS evaluation setting. 
These results show that RippleMem provides stable gains under both dialogue-level and full-history long-term memory evaluation settings.

\paragraph{Larger gains on evidence-distributed questions.}
The improvements are especially clear on question types that require connecting dispersed memories. On LoCoMo temporal questions, RippleMem improves over SimpleMem by 5.45 F1 points and 9.66 judge-accuracy points. On open-domain questions, RippleMem achieves the highest judge accuracy, improving over RF-Mem from 68.75 to 70.83, although RF-Mem obtains higher lexical overlap. LongMemEval-S shows a similar pattern across the two comparison groups: RippleMem improves multi-session reasoning over SimpleMem (78.20 vs. 60.92) and EverMemOS (80.45 vs. 73.68), and remains competitive on temporal and knowledge-update questions where answers often depend on resolving event order or changed user states. This pattern is consistent with RippleMem's design because expanding from recalled anchors helps recover supporting memories that are not directly matched by the original query.

\paragraph{Grounded recollection matters.}
Association- and recollection-style baselines provide competitive memory access signals. RF-Mem achieves the strongest baseline judge score on LoCoMo under
matched memory units and evidence budgets, while REMem, MemGAS, and
M-Flow introduce agentic, multi-granularity, or graph-routed retrieval,
respectively. Nevertheless, RippleMem achieves the best overall judge accuracy and the strongest results in the temporal, open-domain, and single-hop categories. This suggests that the gain comes not merely from adding association or iterative retrieval, but from grounding recollection in stored event memories and using already recalled evidence to recover additional support. Appendix~\ref{app:adversarial} further evaluates this behavior on adversarial no-support questions, where the dialogue provides no valid supporting evidence.

\subsection{Ablation Study}
\label{sec:ablation}

\begin{figure}[!t]
    \centering
    \includegraphics[width=\linewidth]{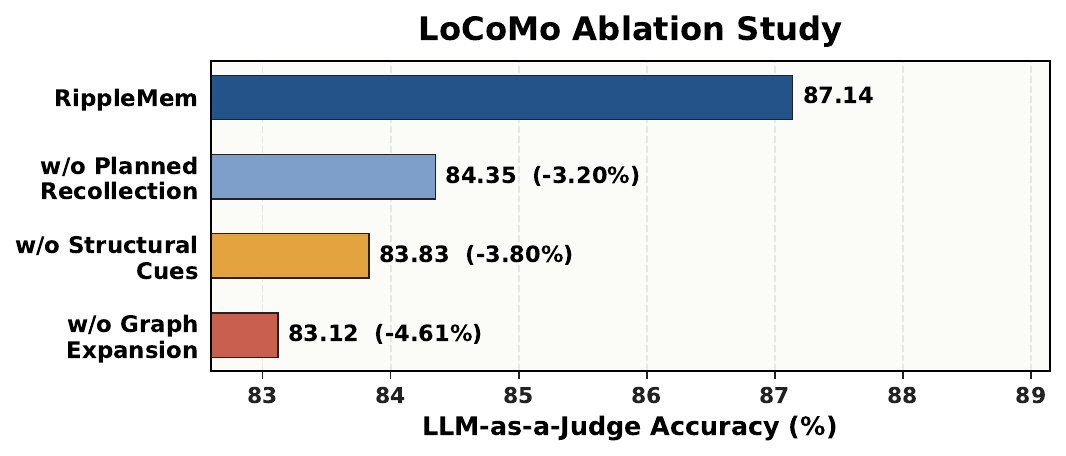}
    \vspace{-22pt}
    \caption{Ablation results on LoCoMo using LLM-as-a-Judge accuracy. Values in parentheses denote relative decreases compared with the full RippleMem model.}
    \label{fig:ablation-locomo}
\end{figure}

We conduct ablations on LoCoMo to isolate the contributions of structural cues, anchor-local graph expansion, and planned recollection. The backbone model, memory extraction, and answer generation are kept fixed; only the corresponding memory-access component is varied. Full ablation results with F1, BLEU-1, and judge accuracy are provided in Appendix~\ref{app:full-ablation}. A separate controlled replacement study of the edge-construction mechanism is reported in Appendix~\ref{app:llm-edge}.

Figure~\ref{fig:ablation-locomo} shows that removing graph expansion causes the largest drop in judge accuracy, from 87.14 to 83.12, indicating that first-hop recall often misses supporting evidence. Removing structural cues lowers accuracy to 83.83, especially when related memories use different wording but share participants, locations, or temporal cues. Planned recollection also helps: without it, accuracy drops to 84.35 despite using the same expansion budget.

\subsection{Phase-Level Efficiency Analysis}
\label{sec:cost-analysis}

\begin{table}[!t]
\centering
\footnotesize
\setlength{\tabcolsep}{2.2pt}
\renewcommand{\arraystretch}{1.10}
\begin{tabular}{@{}lrrrr@{}}
\toprule
Method
& Build (s) $\downarrow$
& Build Tok. $\downarrow$
& Ans. Ctx. $\downarrow$
& J $\uparrow$ \\
\midrule
Mem0$^\mathrm{g}$
& 3623.63 & 4,243,278 & \textbf{628.17} & 65.97 \\
Zep
& 3532.03 & 6,037,130 & 1,629.50 & 76.69 \\
RippleMem
& \textbf{117.51} & \textbf{87,097} & 1,471.93 & \textbf{87.14} \\
\bottomrule
\end{tabular}
\caption{Phase-level cost comparison on LoCoMo. Build time and build tokens are averaged per dialogue; answer-context tokens are averaged per question. J denotes overall LLM-as-a-Judge accuracy.}
\label{tab:phase-cost}
\end{table}

Table~\ref{tab:phase-cost} decomposes efficiency into offline memory construction and answer-time evidence context. Compared with Mem0$^\mathrm{g}$ and Zep, RippleMem reduces construction time by approximately $30\times$ and construction-stage tokens by $48.7$--$69.3\times$. Meanwhile, it maintains a relatively compact average answer-context budget of 1,471.93 tokens per question and achieves the strongest overall judge accuracy. These results demonstrate a favorable phase-level trade-off between persistent-memory construction, answer-time context size, and downstream effectiveness. The additional cost of query-time recollection control is analyzed separately in Appendix~\ref{app:controller-cost}.

\section{Conclusion}

In this paper, we introduced RippleMem, a long-term memory system for LLM agents that shifts memory access from isolated retrieval to adaptive associative recollection. 
Informed by cognitive views of event-based memory organization and cue-dependent recollection, RippleMem represents interaction history as cue-rich episodic memory units, links them in an event-centric memory graph, and recovers missing support through memory-anchor-based recollection. 
Experiments on LoCoMo and LongMemEval-S show strong overall performance, with particularly clear gains on evidence-distributed questions. 
These results suggest that reliable long-term memory depends not only on storing past experience, but also on organizing it so that partial recall can guide further evidence recovery. 
More broadly, recollection-aware memory access offers a promising direction for building more reliable, efficient, and context-sensitive LLM agents.

\section*{Limitations}

RippleMem is evaluated primarily on text-only long-term conversational memory benchmarks. This setting does not cover multimodal interaction, embodied agents, or tool-use environments, where memory may involve visual observations, actions, and external tool states. Extending associative recollection to such settings remains an important direction for future work.

RippleMem uses LLM-mediated operations for memory extraction, query analysis, and recollection planning. While these operations enable flexible memory construction and adaptive recollection, they add latency and cost compared with single-pass retrieval baselines. Some components can be cached, batched, or executed asynchronously, but further improving end-to-end efficiency remains useful for large-scale deployment.

Finally, current benchmarks provide limited stress tests for continuously growing personal memory over very long timelines. Future evaluations with longer histories, evolving user states, memory aging, and privacy-preserving deletion would provide a more complete assessment of long-term agent memory systems.

\section*{Acknowledgments}
    This work is supported by the National Natural Science Foundation of China (No. 62206259).
    
\bibliography{custom}

\appendix
\raggedbottom

\section{Evaluation Protocol and Implementation Details}
\label{app:evaluation}

\subsection{Evaluation Settings}
\label{app:evaluation-settings}

For LoCoMo, we evaluate on 1,540 questions from 10 ultra-long dialogues, covering single-hop, multi-hop, temporal, and open-domain question answering. We report F1, BLEU-1, and LLM-as-a-Judge accuracy.

For LongMemEval-S, we evaluate on 500 questions over long user-agent interaction histories. The benchmark includes question types such as multi-session reasoning, temporal reasoning, knowledge update, user-specific facts, assistant-provided information, and preference recall.

For LongMemEval-S, we report two prior-aligned comparison settings to ensure comparability with previously reported baselines. In the SimpleMem-aligned setting, RippleMem uses GPT-4.1-mini as the LLM judge and Qwen3-Embedding-0.6B as the dense encoder. In the EverMemOS-aligned setting, RippleMem uses GPT-4o-mini as the LLM judge and Qwen3-Embedding-4B as the dense encoder. Baseline scores in each group are taken from the corresponding prior work, while RippleMem is evaluated under the same judge and encoder configuration.

\subsection{Baseline Descriptions}
\label{app:baseline-descriptions}

\textbf{Full-Context.}
This baseline directly provides the available interaction history to the LLM without using an external memory module, serving as a long-context prompting reference.

\textbf{Mem0 and Mem0$^\mathrm{g}$}~\citep{chhikara2025mem0buildingproductionreadyai}.
Mem0 extracts compact long-term memories from interactions and retrieves them for downstream answering. Mem0$^\mathrm{g}$ augments this design with graph memory to capture relations among stored memories.

\textbf{Zep}~\citep{rasmussen2025zeptemporalknowledgegraph}.
Zep maintains a temporal knowledge graph over user interactions, supporting memory retrieval through structured entity and relation updates over time.

\textbf{MemGAS}~\citep{xu2025singlemultigranularitylongtermmemory}.
MemGAS constructs memory associations across multiple granularities and selects relevant memory units for long-term conversational reasoning.

\textbf{M-Flow}~\citep{mflow2026github}.
M-Flow organizes memory through a flow-style retrieval process, using structured memory routing to support multi-step access over stored interaction records.

\textbf{REMem}~\citep{shu2026remem}.
REMem constructs a hybrid episodic memory graph from time-aware gists
and time-scoped facts, and uses an agentic retriever with semantic,
lexical, temporal, and graph-exploration tools for iterative evidence
gathering.

\textbf{SimpleMem}~\citep{liu2026simplememefficientlifelongmemory}.
SimpleMem improves lifelong memory through semantic compression, structured indexing, and query-aware retrieval planning over compact memory entries.

\textbf{RF-Mem}~\citep{zhang2026evokingusermemorypersonalizing}.
RF-Mem proposes a familiarity--recollection retrieval mechanism that adapts between direct memory retrieval and recollection-style expansion. In our LoCoMo comparison, we evaluate its retrieval-side mechanism over the same extracted memory units as RippleMem and match the per-question evidence budget, isolating the effect of memory-access strategy. This setting does not reuse RippleMem's graph expansion or recollection controller.

\textbf{LightMem}~\citep{fang2026lightmemlightweightefficientmemoryaugmented}.
LightMem focuses on lightweight memory management for efficient long-term memory use, reducing storage and retrieval overhead while maintaining user-specific information.

\textbf{MemU}~\citep{memu2025github}.
MemU is a memory management framework for LLM agents that maintains user memory through persistent storage and retrieval-oriented updates.

\textbf{MemOS}~\citep{li2025memosmemoryosai}.
MemOS treats memory as a managed system resource and provides a unified framework for scheduling, storing, and retrieving different types of memory.

\textbf{EverMemOS}~\citep{hu2026evermemosselforganizingmemoryoperating}.
EverMemOS models memory as a lifecycle that transforms episodic traces into consolidated memory structures and uses reconstructive recollection for long-horizon reasoning.

\subsection{Metric Aggregation}
\label{app:metric-aggregation}

For LoCoMo, category-level scores are reported for diagnostic analysis, while average scores are computed over all evaluated questions rather than by macro-averaging category-level scores. For LongMemEval-S, overall accuracy is computed as the total number of correctly judged answers divided by the total number of questions.

\subsection{LLM-as-a-Judge Protocol}
\label{app:judge-protocol}

We use an LLM-as-a-Judge protocol to evaluate semantic correctness when lexical overlap is insufficient for long-form memory answers. The judge receives only the question, the reference answer, and the generated answer, and returns a binary correctness label. The same judge model is used for all methods within each evaluation setting, with temperature set to zero. The judge prompt is provided in Appendix~\ref{app:prompts}. For LoCoMo, GPT-4.1-mini is used as the judge. For LongMemEval-S, the judge follows the corresponding prior-aligned setting described in Appendix~\ref{app:evaluation-settings}.

\subsection{Implementation Settings}
\label{app:implementation-settings}

Table~\ref{tab:implementation-settings} reports the benchmark-specific settings used by RippleMem. All values are fixed before evaluation and are not tuned per question or per category. The round budget $R$ counts additional anchor-local recollection rounds after first-hop recall; we set $R=1$ for all reported experiments.

\begin{table}[H]
\centering
\footnotesize
\setlength{\tabcolsep}{3pt}
\renewcommand{\arraystretch}{1.05}
\begin{tabular}{@{}lcc@{}}
\toprule
Setting & LoCoMo & LongMemEval-S \\
\midrule
Memory window & 40 turns & 10 turns \\
Window overlap & 2 turns & 0 turns \\
Semantic recall top-$k$ & 10 & 15 \\
Lexical recall top-$k$ & 5 & 8 \\
Structured recall top-$k$ & 5 & 8 \\
Expansion hops & 2 & 2 \\
Max anchors & 3 & 3 \\
Semantic expansion top-$k$ & 5 & 5 \\
Structural expansion top-$k$ & 5 & 5 \\
Final evidence budget & 30 & 30 \\
Additional recollection rounds $R$ & 1 & 1 \\
\bottomrule
\end{tabular}
\caption{Benchmark-specific implementation settings for RippleMem.}
\label{tab:implementation-settings}
\end{table}

Table~\ref{tab:graph-settings} reports the graph construction settings used by RippleMem. These values are also fixed before evaluation.

\begin{table}[H]
\centering
\small
\begin{tabular}{lc}
\toprule
Setting & Value \\
\midrule
Semantic edge candidate pool & 20 \\
Cue-based edge candidate pool & 20 \\
Max semantic edges per node & 6 \\
Max structural edges per node & 6 \\
Semantic edge threshold & 0.85 \\
Structural edge threshold & 0.60 \\
Participant cue weight & 0.50 \\
Location cue weight & 0.20 \\
Temporal cue weight & 0.30 \\
Temporal decay scale & 7 days \\
\bottomrule
\end{tabular}
\caption{Graph construction settings used by RippleMem.}
\label{tab:graph-settings}
\end{table}

\paragraph{Source-aware ordering weights.}
The evidence ordering score uses fixed weights across all experiments. 
We set $\lambda_q=1.25$, $\lambda_p=1.0$, and $\lambda_a=0.05$ for all experiments. 
The provenance term $\pi(m)$ aggregates retrieval-source contributions with fixed source weights: semantic search $1.0$, lexical search $0.75$, cue-based lookup $0.9$, and anchor-local expansion $0.85$. 
For each source, rank support is computed as $1/(5+\mathrm{rank})$ and combined with the available normalized source score. 
These weights are fixed before evaluation and are not tuned per dataset, category, or test question.

\section{Additional Experimental Results}
\label{app:additional-results}

\subsection{Aligned Comparison with MemPalace}
\label{app:mempalace}

MemPalace~\citep{mempalace2026github} reports session-level retrieval
recall, which is not directly comparable to end-to-end QA accuracy.
We therefore conduct an aligned end-to-end comparison using the same
LoCoMo split, Qwen3-Embedding-0.6B encoder, GPT-4.1-mini answer model
and judge, and answer-level evaluation procedure. We retain
MemPalace's hybrid-v5, session-level retrieval without LLM reranking,
and evaluate both Top-5 and Top-10 configurations.

\begin{table}[!t]
\centering
\small
\setlength{\tabcolsep}{7pt}
\begin{tabular}{lrr}
\toprule
Method & J $\uparrow$ & Answer Ctx. $\downarrow$ \\
\midrule
MemPalace (Top-5)  & 81.75 & 3,802.01 \\
MemPalace (Top-10) & 86.69 & 7,502.59 \\
RippleMem           & \textbf{87.14} & \textbf{1,471.93} \\
\bottomrule
\end{tabular}
\caption{
Aligned end-to-end comparison on LoCoMo Categories 1--4.
J denotes LLM-as-a-Judge accuracy, and answer-context tokens are
averaged per question. Top-10 is MemPalace's documented LoCoMo
configuration.
}
\label{tab:mempalace-aligned}
\end{table}

Compared with MemPalace Top-5, RippleMem improves judge accuracy by
5.39 points while using approximately $2.6\times$ fewer answer-context
tokens. Under MemPalace's documented Top-10 configuration, RippleMem
achieves slightly higher judge accuracy (87.14 versus 86.69) while
using approximately $5.1\times$ fewer answer-context tokens, an
80.4\% reduction. This result shows that RippleMem maintains higher
end-to-end answer accuracy with substantially more compact evidence,
rather than relying on a larger answer context.

\subsection{Evaluation on EverMemBench}
\label{app:evermembench}

\paragraph{Benchmark.}
We additionally evaluate RippleMem on EverMemBench, using its
EverMemBench-Dynamic release~\citep{hu2026evermembench}.
Unlike the dyadic conversations in LoCoMo and LongMemEval-S,
EverMemBench evaluates long-term memory in multi-party, multi-group
conversations with cross-group interactions, evolving information,
and role-specific personas. It contains five project histories with
51,023 dialogue turns and 2,400 questions, covering three dimensions
and nine subtasks: fine-grained recall (single-hop, multi-hop, and
temporal), memory awareness (constraint, proactivity, and update),
and profile understanding (style, skill, and role).

\paragraph{Experimental setup.}
We compare RippleMem with Full Context, MemoBase, Mem0, Zep, MemOS,
and RF-Mem. Full Context receives the complete dialogue history,
following the EverMemBench setting. MemoBase, Mem0, Zep, and MemOS
follow the implementations and memory/retrieval settings reported by
the benchmark. RF-Mem follows the controlled setup used in our main
experiments: it reuses RippleMem's memory units and embeddings, with
the number of returned memories matched to RippleMem for each
question. RippleMem uses its event-centric memory graph construction
and associative recollection pipeline.

All methods are evaluated on the same benchmark split using
GPT-4.1-mini for answer generation and for judging open-ended
questions, with the same scoring pipeline. Multiple-choice questions
are evaluated by exact option matching. For RippleMem, we use BGE-M3
as the dense encoder, a memory window of 20 turns with one-turn
overlap, semantic/lexical/structured retrieval budgets of 10/5/5,
and a final evidence budget of 30 memories. These settings are fixed
across all questions and subtasks. Overall accuracy is computed over
all 2,400 questions rather than by macro-averaging the nine subtask
scores.

\begin{table*}[t]
\centering
\footnotesize
\setlength{\tabcolsep}{3.8pt}
\renewcommand{\arraystretch}{1.10}
\resizebox{\textwidth}{!}{
\begin{tabular}{l|ccc|ccc|ccc|c}
\toprule
\multirow{2}{*}{Method}
& \multicolumn{3}{c|}{Fine-Grained Recall}
& \multicolumn{3}{c|}{Memory Awareness}
& \multicolumn{3}{c|}{Profile Understanding}
& \multirow{2}{*}{Overall} \\
& Single & Multi & Temp.
& Const. & Proact. & Update
& Style & Skill & Role & \\
\midrule
Full Context
& 84.18 & 1.62 & 9.14
& 62.91 & 27.83 & 40.31
& \textbf{39.76} & 34.92 & 38.85
& 37.58 \\

MemoBase
& 58.77 & 15.36 & 20.72
& 63.94 & 34.02 & 31.44
& 15.18 & 30.22 & 39.64
& 34.48 \\

Mem0
& 56.08 & 9.11 & 4.56
& 67.03 & 55.74 & 50.92
& 24.11 & 30.77 & 35.48
& 37.23 \\

Zep
& 71.62 & 10.47 & 15.38
& 66.43 & 45.18 & 44.92
& 28.64 & 34.81 & 45.12
& 40.16 \\

MemOS
& 72.14 & 21.76 & 18.35
& 68.64 & 54.82 & 42.09
& 30.61 & 31.82 & 47.36
& 42.73 \\

RF-Mem
& 91.55 & 15.66 & 21.00
& 77.36 & 67.21 & 54.10
& 30.68 & \textbf{39.64} & 49.49
& 52.42 \\

\textbf{RippleMem}
& \textbf{92.02} & \textbf{22.09} & \textbf{21.33}
& \textbf{78.11} & \textbf{71.66} & \textbf{58.96}
& 31.25 & 36.69 & \textbf{53.06}
& \textbf{54.75} \\
\bottomrule
\end{tabular}}
\caption{Results on EverMemBench. All values are accuracy percentages.
Best results in each column are bolded. Overall accuracy is computed
over all 2,400 questions.}
\label{tab:evermembench}
\end{table*}

\paragraph{Results.}
RippleMem achieves the highest overall accuracy of 54.75\%,
representing relative improvements of 28.13\% over MemOS and 4.44\%
over the budget-matched RF-Mem baseline. It performs best on seven of
the nine subtasks: single-hop, multi-hop, temporal, constraint,
proactivity, update, and role. The gains therefore extend beyond
factual retrieval to questions requiring cross-group evidence
integration, evolving-memory resolution, and role-sensitive
reasoning. Its improvement over RF-Mem under matched memory units and
evidence budgets further indicates that evidence-conditioned
associative recollection contributes beyond the use of
recollection-style retrieval alone.

\subsection{Full Ablation Results}
\label{app:full-ablation}

Table~\ref{tab:full-ablation} reports the complete LoCoMo ablation results with F1, BLEU-1, and LLM-as-a-Judge accuracy. The main paper visualizes judge accuracy for readability, while this table provides the full metric breakdown.

\begin{table*}[t]
\centering
\footnotesize
\setlength{\tabcolsep}{4.0pt}
\renewcommand{\arraystretch}{1.12}
\resizebox{\textwidth}{!}{
\begin{tabular}{l|ccc|ccc|ccc|ccc|ccc}
\toprule
\multirow{2}{*}{Method}
& \multicolumn{3}{c|}{Multi-Hop}
& \multicolumn{3}{c|}{Temporal}
& \multicolumn{3}{c|}{Open Domain}
& \multicolumn{3}{c|}{Single Hop}
& \multicolumn{3}{c}{Average} \\
& F1 $\uparrow$ & B1 $\uparrow$ & J $\uparrow$
& F1 $\uparrow$ & B1 $\uparrow$ & J $\uparrow$
& F1 $\uparrow$ & B1 $\uparrow$ & J $\uparrow$
& F1 $\uparrow$ & B1 $\uparrow$ & J $\uparrow$
& F1 $\uparrow$ & B1 $\uparrow$ & J $\uparrow$ \\
\midrule
w/o Structural Cues
& 37.24 & 28.90 & 74.82
& 59.96 & 45.85 & 81.93
& \textbf{29.98} & \textbf{24.79} & 62.50
& 55.37 & 49.17 & 90.01
& 51.43 & 43.24 & 83.83 \\
w/o Graph Expansion
& 37.42 & 28.99 & 72.70
& 58.81 & 44.29 & 82.87
& 25.98 & 21.40 & 61.46
& 54.87 & 48.62 & 89.18
& 50.69 & 42.43 & 83.12 \\
w/o Planned Recollection
& 36.36 & 28.90 & 77.66
& 60.37 & 45.95 & 81.31
& 26.39 & 22.28 & 60.42
& 55.20 & 48.94 & 90.49
& 51.03 & 42.99 & 84.35 \\
RippleMem
& \textbf{38.58} & \textbf{31.12} & \textbf{77.67}
& \textbf{62.37} & \textbf{47.29} & \textbf{85.67}
& 25.71 & 20.93 & \textbf{70.83}
& \textbf{56.44} & \textbf{49.79} & \textbf{92.75}
& \textbf{52.49} & \textbf{44.05} & \textbf{87.14} \\
\bottomrule
\end{tabular}
}
\caption{Full ablation results on LoCoMo. We report F1, BLEU-1 (B1), and LLM-as-a-Judge accuracy (J). Bold denotes the best result for each metric.}
\label{tab:full-ablation}
\end{table*}

A small discrepancy appears between lexical-overlap metrics and judge accuracy. For example, w/o Structural Cues obtains higher F1/BLEU-1 in open-domain questions, but its judge accuracy is lower than the full model. This suggests that semantic-only expansion can sometimes retrieve memories with overlapping wording, while structural associations help organize grounded cues into evidence that better supports semantically correct answers.

\subsection{Edge Construction Mechanism Ablation}
\label{app:llm-edge}

We conduct a replacement ablation to examine whether RippleMem's typed association scoring can be replaced by LLM-based edge decisions. We construct a controlled variant, denoted \textsc{LLM-Edge}, in which only the edge-decision mechanism is replaced by GPT-4.1-mini, while all other components remain unchanged. The extracted memory units, candidate pool, per-node degree budget, recollection controller, graph expansion, answer generator, and judge remain unchanged. Both variants consider the same union of semantic top-20 and cue-based top-20 candidates. For each candidate pair, the verifier predicts \textsc{Sem-Supports}, \textsc{Struct-Supports}, \textsc{Both}, or \textsc{None}. Lexical matching remains a query-time recall channel and does not create persistent graph edges in either variant.

Table~\ref{tab:llm-edge-performance} reports the effectiveness comparison. 
The LLM-based verifier improves multi-hop judge accuracy from 77.67 to 79.79, indicating that flexible edge judgments can recover useful relations in some cases. 
However, it does not improve overall performance, obtaining 86.62 judge accuracy compared with 87.14 for RippleMem.

\begin{table*}[!t]
\centering
\footnotesize
\setlength{\tabcolsep}{4.0pt}
\renewcommand{\arraystretch}{1.12}
\resizebox{\textwidth}{!}{
\begin{tabular}{l|ccc|ccc|ccc|ccc|ccc}
\toprule
\multirow{2}{*}{Method}
& \multicolumn{3}{c|}{Multi-Hop}
& \multicolumn{3}{c|}{Temporal}
& \multicolumn{3}{c|}{Open Domain}
& \multicolumn{3}{c|}{Single Hop}
& \multicolumn{3}{c}{Average} \\
& F1 $\uparrow$ & B1 $\uparrow$ & J $\uparrow$
& F1 $\uparrow$ & B1 $\uparrow$ & J $\uparrow$
& F1 $\uparrow$ & B1 $\uparrow$ & J $\uparrow$
& F1 $\uparrow$ & B1 $\uparrow$ & J $\uparrow$
& F1 $\uparrow$ & B1 $\uparrow$ & J $\uparrow$ \\
\midrule
\textsc{LLM-Edge}
& 37.81 & 30.65 & \textbf{79.79}
& 61.67 & 46.69 & 83.80
& \textbf{30.31} & \textbf{24.66} & \textbf{70.83}
& 55.64 & 48.98 & 91.80
& 52.05 & 43.63 & 86.62 \\
RippleMem
& \textbf{38.58} & \textbf{31.12} & 77.67
& \textbf{62.37} & \textbf{47.29} & \textbf{85.67}
& 25.71 & 20.93 & \textbf{70.83}
& \textbf{56.44} & \textbf{49.79} & \textbf{92.75}
& \textbf{52.49} & \textbf{44.05} & \textbf{87.14} \\
\bottomrule
\end{tabular}
}
\caption{
Effect of replacing RippleMem's typed association scoring with GPT-4.1-mini edge decisions on LoCoMo.
F1, B1, and J denote F1, BLEU-1, and LLM-as-a-Judge accuracy.
Average scores are computed over all evaluated questions.
Bold denotes the better result in each column.
}
\label{tab:llm-edge-performance}
\end{table*}

Table~\ref{tab:llm-edge-cost} compares the corresponding offline construction costs. 
\textsc{LLM-Edge} requires approximately $27.7\times$ more construction tokens and $33.0\times$ more construction calls. 
RippleMem's construction calls arise from memory extraction, whereas \textsc{LLM-Edge} additionally invokes one batched candidate-edge verification call per source memory. 
Thus, although LLM-mediated edge decisions benefit particular question types, RippleMem's typed associations provide a stronger overall accuracy--construction-cost trade-off.

\begin{table}[!t]
\centering
\small
\setlength{\tabcolsep}{5pt}
\renewcommand{\arraystretch}{1.1}
\begin{tabular*}{\columnwidth}{@{\extracolsep{\fill}}lrr}
\toprule
Method
& Build Tokens $\downarrow$
& Build Calls $\downarrow$ \\
\midrule
\textsc{LLM-Edge} & 2,408,796 & 521.2 \\
RippleMem & \textbf{87,097} & \textbf{15.8} \\
\bottomrule
\end{tabular*}
\caption{
Offline construction cost of the two edge-decision mechanisms on LoCoMo.
Values are averaged per dialogue; lower is better.
}
\label{tab:llm-edge-cost}
\end{table}

\subsection{Query-Time Recollection Controller Cost}
\label{app:controller-cost}

Once query cues have been extracted, semantic, lexical, and cue-based lookup and bounded graph traversal require no additional LLM calls. RippleMem additionally invokes a schema-constrained recollection controller to assess evidence sufficiency, select memory anchors, and construct a missing-support target. On LoCoMo, this controller consumes an average of 2,880.6 prompt and completion tokens per question. Since the round budget is fixed to $R=1$, the controller is invoked at most once after first-hop recall. The corresponding ablation shows its utility: removing planned recollection under the same expansion budget reduces judge accuracy from 87.14 to 84.35.

\subsection{Adversarial No-Support Analysis}
\label{app:adversarial}

LoCoMo Category 5 contains 446 adversarial questions for which the dialogue provides no valid answer support. Correct handling therefore requires recognizing insufficient evidence rather than producing an unsupported answer. RippleMem follows a conservative strategy: rather than immediately declaring that no support exists, the controller may invoke one additional bounded verification round ($R=1$). Anchor-local expansion returns an empty support set when no candidate matches the missing-support target, so the evidence state is updated only when suitable support is found.

\begin{table}[!t]
\centering
\small
\setlength{\tabcolsep}{5pt}
\renewcommand{\arraystretch}{1.08}
\begin{tabular*}{\columnwidth}{@{\extracolsep{\fill}}lr}
\toprule
Method & Category-5 J (\%) \\
\midrule
Zep & 35.43 \\
Mem0$^\mathrm{g}$ & 38.12 \\
% MemPalace & 72.65 \\
SimpleMem & 86.32 \\
RF-Mem$^\dagger$ & \textbf{88.57} \\
RippleMem w/o Round 2 & 84.75 \\
RippleMem & \emph{86.32} \\
\bottomrule
\end{tabular*}
\caption{
LLM-as-a-Judge accuracy on LoCoMo Category-5 adversarial no-support questions.
}
\label{tab:category5-accuracy}
\end{table}

\begin{table}[!t]
\centering
\small
\setlength{\tabcolsep}{5pt}
\renewcommand{\arraystretch}{1.08}
\begin{tabular*}{\columnwidth}{@{\extracolsep{\fill}}lrr}
\toprule
Metric
& Categories 1--4
& Category 5 \\
\midrule
Number of questions ($N$) & 1540 & 446 \\
Round-2 trigger rate (R2, \%) & 60.8 & 93.0 \\
J(all) & 87.1 & 86.3 \\
J(R2) & 81.7 & 87.5 \\
J(stop) & 95.5 & 71.0 \\
\bottomrule
\end{tabular*}
\caption{
Recollection-controller behavior on answerable and adversarial no-support LoCoMo questions.
R2 is the percentage of questions that trigger the additional bounded recollection round; J(all), J(R2), and J(stop) denote judge accuracy over all questions, triggered cases, and first-hop stopping cases, respectively.
}
\label{tab:controller-behavior}
\end{table}

Table~\ref{tab:category5-accuracy} shows that RippleMem achieves 86.32\% judge accuracy on adversarial no-support questions. Removing the additional recollection round reduces accuracy to 84.75\%, indicating that bounded verification remains useful even when no valid answer support exists.

Table~\ref{tab:controller-behavior} further shows that the controller triggers an additional round for 93.0\% of Category-5 questions, compared with 60.8\% for Categories 1--4. Accuracy among the triggered adversarial cases reaches 87.5\%. This behavior reflects RippleMem's conservative response to insufficient first-hop evidence: it performs one bounded verification round but updates the evidence state only when suitable support is found.

\section{Reproducibility Artifacts}
\label{app:reproducibility}

\subsection{RippleMem Inference Procedure}
\label{app:inference-procedure}

Algorithm~\ref{alg:ripplemem-inference} summarizes the inference procedure of RippleMem. The algorithm abstracts away implementation-specific indexing details and highlights the controlled recollection loop used at test time.

\begin{algorithm}[!t]
\small
\caption{RippleMem inference with adaptive associative recollection}
\label{alg:ripplemem-inference}
\begin{algorithmic}[1]
\Require query $q$, memory graph $G=(\mathcal{M},\mathcal{E}_{\mathrm{sem}}\cup\mathcal{E}_{\mathrm{str}})$, hop limit $h$, maximum rounds $R$, evidence budget $K$
\State $c_q \gets \mathrm{ExtractQueryCues}(q)$
\State $C_{\mathrm{sem}} \gets \mathrm{SemanticRecall}(q,c_q,\mathcal{M})$
\State $C_{\mathrm{lex}} \gets \mathrm{LexicalRecall}(q,c_q,\mathcal{M})$
\State $C_{\mathrm{cue}} \gets \mathrm{StructuredCueRecall}(c_q,\mathcal{M})$
\State $C \gets C_{\mathrm{sem}} \cup C_{\mathrm{lex}} \cup C_{\mathrm{cue}}$
\For{$r=1$ to $R$}
    \State $(d_r,A_r,g_r,s_r) \gets \Pi_{\mathrm{rec}}(q,c_q,C)$
    \If{$d_r=\mathrm{STOP}$ or $A_r=\emptyset$}
        \State \textbf{break}
    \EndIf
    \State $U^{\mathrm{sem}}_r \gets \mathrm{Expand}(G_{\mathrm{sem}},A_r,h)\setminus C$
    \State $U^{\mathrm{str}}_r \gets \mathrm{Expand}(G_{\mathrm{str}},A_r,h)\setminus C$
    \State $S^{\mathrm{sem}}_r \gets \mathrm{SelectTop}(U^{\mathrm{sem}}_r,g_r)$
    \State $S^{\mathrm{str}}_r \gets \mathrm{SelectTop}(U^{\mathrm{str}}_r,g_r)$
    \State $S_r \gets \mathrm{Dedup}(S^{\mathrm{sem}}_r \cup S^{\mathrm{str}}_r)$
    \If{$S_r=\emptyset$}
        \State \textbf{break}
    \EndIf
    \State $C \gets C \cup S_r$
\EndFor
\State $\bar{C} \gets \mathrm{ConsolidateByMemoryId}(C)$
\State $E_K \gets \mathrm{TopK}_{\rho}(\bar{C},K)$
\State \Return $E_K$
\end{algorithmic}
\end{algorithm}

\section{Qualitative Analysis}
\label{app:qualitative}

\subsection{Case Studies}
\label{app:case-study}

\begin{figure*}[t]
    \centering
    \includegraphics[width=0.95\textwidth]{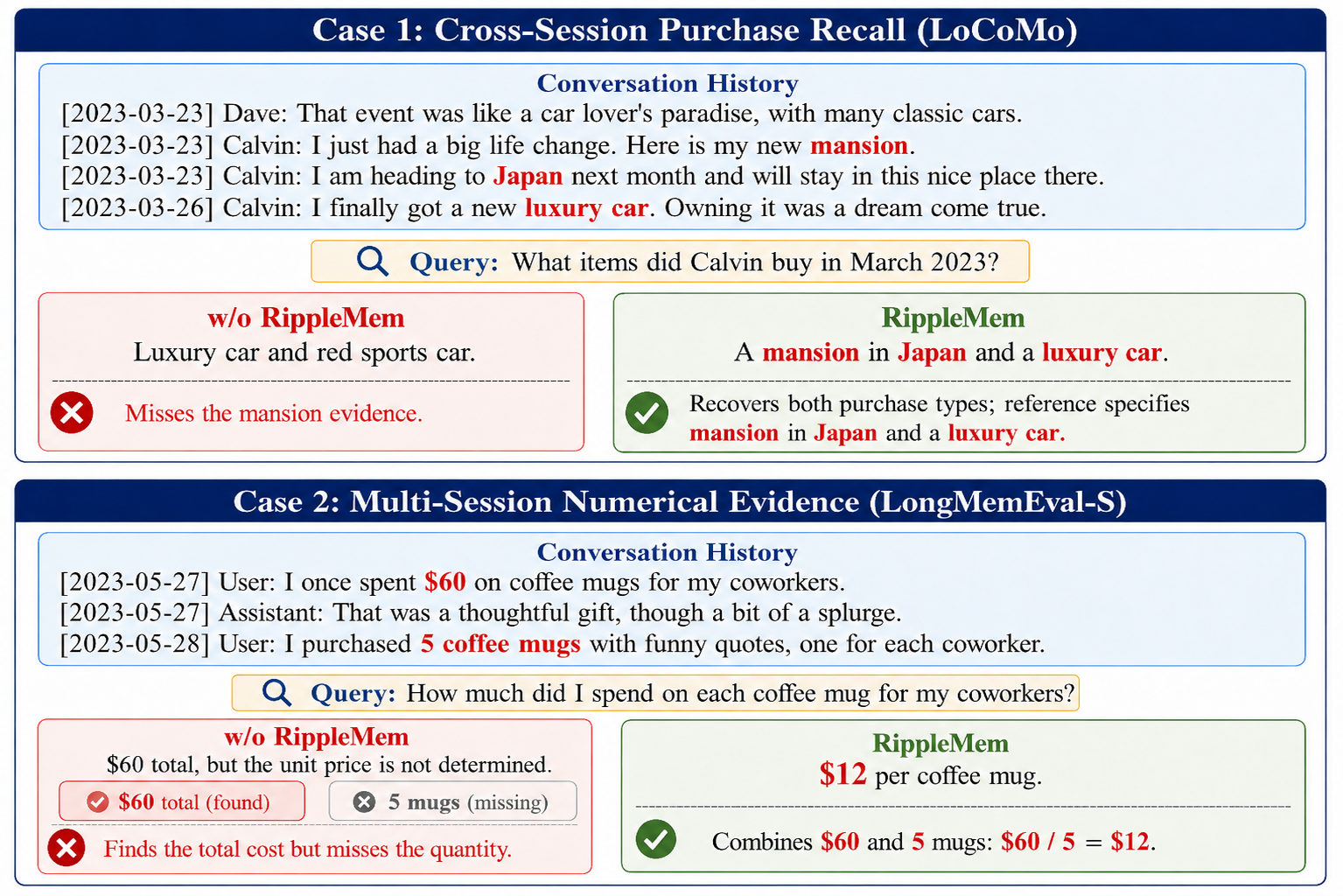}
    \caption{
    Case studies on LoCoMo and LongMemEval-S.
    In both examples, the answer depends on evidence distributed across multiple conversation snippets.
    RippleMem recovers and composes the missing support, while the comparison setting produces an incomplete answer from partial evidence.
    }
    \label{fig:case-study}
\end{figure*}

Figure~\ref{fig:case-study} illustrates how RippleMem handles queries whose answers require evidence composition across sessions. 
The LoCoMo example requires recovering multiple purchase-related memories, while the LongMemEval-S example requires combining a total cost with a separately mentioned quantity. 
In both cases, the comparison setting answers from partial evidence, whereas RippleMem recovers the missing support needed to produce an answerable evidence context.

\subsection{Worked Write-to-Read Trace}
\label{app:write-read-trace}

Table~\ref{tab:write-read-trace} presents a complete write-to-read replay that connects event-graph construction with query-time associative recollection. The example shows how typed associations established during the write phase enable RippleMem to recover recommendation evidence missing from first-hop recall.

\begin{table*}[t]
\centering
\footnotesize
\setlength{\tabcolsep}{5pt}
\renewcommand{\arraystretch}{1.14}
\begin{tabular}{p{0.16\textwidth}p{0.78\textwidth}}
\toprule
\textbf{Stage} & \textbf{Trace} \\
\midrule

Query
&
Which books has John recommended to James?
\\

Reference
&
\emph{The Name of the Wind}, \emph{The Stormlight Archive},
\emph{Kingkiller Chronicle}, and \emph{The Expanse}.
\\

Failure example
&
``\emph{The Name of the Wind} fantasy novel trilogy,''
 which omits the other three recommendations.
\\

\midrule
\multicolumn{2}{l}{\textit{Write phase: event-memory construction and association}} \\
\midrule

Event memory $m_1$
&
Source D8:14. John recommended \emph{The Name of the Wind} to James and praised its writing.
Participants: $\{\text{John},\text{James}\}$;
location: $\emptyset$;
time: $[2022\text{-}04\text{-}29\mathrm{T}14{:}36{:}00,\,
2022\text{-}04\text{-}29\mathrm{T}14{:}36{:}00]$.
\\

Event memory $m_2$
&
Source D14:9. James asked John which book series he loves and would recommend.
Participants: $\{\text{James},\text{John}\}$;
location: $\emptyset$;
time: $[2022\text{-}06\text{-}16\mathrm{T}17{:}07{:}00,\,
2022\text{-}06\text{-}16\mathrm{T}17{:}07{:}00]$.
\\

Event memory $m_3$
&
Source D14:10. John recommended \emph{The Stormlight Archive},
\emph{Kingkiller Chronicle}, and \emph{The Expanse}.
Participants: $\{\text{John}\}$;
location: $\emptyset$;
time: $[2022\text{-}06\text{-}16\mathrm{T}17{:}07{:}00,\,
2022\text{-}06\text{-}16\mathrm{T}17{:}07{:}00]$.
\\

Typed associations
&
RippleMem retains both semantic and structural associations between $m_2$ and $m_3$.
Their restatements concern the same recommendation exchange, providing semantic compatibility.
The shared participant John and overlapping time intervals provide structural compatibility.
Consequently, $m_3$ is reachable from $m_2$ through bounded local expansion.
\\

\midrule
\multicolumn{2}{l}{\textit{Read phase: adaptive associative recollection}} \\
\midrule

First-hop evidence $C_0$
&
Hybrid recall retrieves $m_1$, which supplies \emph{The Name of the Wind}, and $m_2$, which identifies the later recommendation exchange.
It also retrieves a follow-up memory in which James thanks John for the recommendations and asks why they are his favorites.
The remaining book-series names are absent from $C_0$.
\\

Recollection anchors $A_r$
&
The controller selects the recommendation-request memory $m_2$ and the recalled follow-up memory as anchors.
These memories locate the relevant exchange but do not yet provide the missing recommendation names.
\\

Target goal
&
Find additional books recommended by John to James.
\\

Target restatement
&
Other books or book series recommended by John to James besides
\emph{The Name of the Wind} trilogy.
\\

Target cues
&
{\raggedright
Participants: $\{\text{John}, \text{James}\}$;
locations: $\emptyset$.\par
Time range: \texttt{2022-06-16T00:00:00} to
\texttt{2022-06-16T23:59:59}.\par}
\\

Anchor-local expansion
&
RippleMem expands from $A_r$ over the retained semantic and structural associations.
The target restatement guides semantic matching, while the participant and temporal cues guide structural matching.
Candidates from both channels are merged by memory identity.
\\

Recovered support $S_r$
&
The second recollection round recovers two supporting memories:
(1) John suggested \emph{The Expanse} for science-fiction fans and described it as epic;
and (2) John recommended \emph{The Stormlight Archive} and
\emph{Kingkiller Chronicle} as favorites, while also suggesting
\emph{The Expanse}.
Together, these memories supply the three recommendation names missing from $C_0$.
\\

Evidence update
&
RippleMem updates the evidence state as $C \leftarrow C \cup S_r$.
Evidence assembly then retains support for all four reference items in the final evidence context $E_K$.
\\

Final answer
&
RippleMem answers:
``\emph{The Name of the Wind}, \emph{The Stormlight Archive},
\emph{Kingkiller Chronicle}, and \emph{The Expanse} series.''
\\

\bottomrule
\end{tabular}
\caption{Worked write-to-read trace for associative recollection. The write phase constructs cue-rich event memories and typed associations; at inference time, first-hop memories identify a local recollection region from which RippleMem recovers recommendation evidence missing from the initial evidence state.}
\label{tab:write-read-trace}
\end{table*}

\section{Prompt Templates}
\label{app:prompts}

This section provides the core prompt templates used by RippleMem. The prompts are shown in their implementation-oriented form for reproducibility. For benchmarks with anonymous user/assistant roles, speaker-dependent references may be mapped to fixed canonical role names before cue extraction, so that role information can be represented in the participant field without changing the query meaning.
Placeholders such as \texttt{\{query\}}, \texttt{\{dialogue\_text\}}, and \texttt{\{current\_evidence\}} are filled at runtime. Table~\ref{tab:prompt-overview} summarizes the purpose and output format of each prompt. The implementation field \texttt{lossless\_restatement} corresponds to the memory restatement $r_i$ in Section~\ref{sec:memory-construction}.

\onecolumn
\begin{table}[H]
\centering
\small
\begin{tabular}{lll}
\toprule
Prompt & Purpose & Output \\
\midrule
Memory extraction & Construct cue-rich memory units & JSON memories \\
Query cue extraction & Parse first-hop retrieval cues & Query/cue JSON \\
Recollection planning & Select anchors and target & Plan JSON \\
Answer generation & Generate grounded response & Answer JSON \\
LLM-as-a-Judge & Evaluate correctness & Binary label \\
\bottomrule
\end{tabular}
\caption{Overview of prompt templates used by RippleMem.}
\label{tab:prompt-overview}
\end{table}

\Needspace{0.80\textheight}
\begin{PromptTemplate}{Memory Extraction Prompt}
Your task is to extract all valuable information from the following dialogues and convert them into structured memory entries.

Dialogue IDs in this window: {dialogue_ids}

[Current Window Dialogues]
{dialogue_text}

[Requirements]
1. Complete Coverage: Generate enough memory entries to ensure ALL valuable information in the dialogues is captured.
2. Force Disambiguation: Do not use pronouns such as he, she, it, they, this, or that.
3. Lossless Information: Each entry's lossless_restatement must be a complete, independent, understandable sentence.
4. Precise Extraction:
   - time_range: a pair [start, end] in ISO 8601 format if the memory can be grounded to an absolute time or interval from dialogue context; otherwise null.
   - If the memory refers to a specific event time, represent it as [t, t].
   - If the memory refers to a day, week, month, year, or other interval, represent the full interval.
   - If the dialogue only gives a vague time that cannot be grounded to an absolute interval, keep that wording in lossless_restatement and set time_range to null.
   - locations: all specific location names explicitly grounded in this memory.
   - persons: all explicit human personal names mentioned in this memory. Only include actual individual people.

[Output Format]
Return JSON only:
{
  "memories": [
    {
      "lossless_restatement": "Complete unambiguous restatement.",
      "time_range": ["ISO_START", "ISO_END"] or null,
      "locations": ["location1", "location2"],
      "persons": ["name1", "name2"]
    }
  ]
}

[Good Examples]
Example 1
Dialogues:
[2025-11-15T14:30:00] Alice: Bob, let's meet at Starbucks tomorrow at 2pm to discuss the new product
[2025-11-15T14:31:00] Bob: Okay, I'll prepare the materials

Output:
{
  "memories": [
    {
      "lossless_restatement": "Alice suggested at 2025-11-15T14:30:00 to meet with Bob at Starbucks on 2025-11-16T14:00:00 to discuss the new product.",
      "time_range": ["2025-11-16T14:00:00", "2025-11-16T14:00:00"],
      "locations": ["Starbucks"],
      "persons": ["Alice", "Bob"]
    },
    {
      "lossless_restatement": "Bob agreed to attend the meeting and committed to prepare relevant materials.",
      "time_range": null,
      "locations": [],
      "persons": ["Bob"]
    }
  ]
}

Now process the above dialogues. Return ONLY the JSON object, no other explanations.
\end{PromptTemplate}

\Needspace{0.48\textheight}
\begin{PromptTemplate}{First-Hop Query Cue Extraction Prompt}
You are extracting retrieval cues for first-hop memory retrieval.

Given the question, produce a single JSON object with:
- original_query: the original question
- semantic_query: the full original query with only speaker references normalized; replace first-person references such as I, me, and my with Alice, and replace references to the assistant with Bob when applicable. Do not delete any content or information.
- keywords: 2-6 short lexical cues for BM25/fulltext retrieval
- persons: explicit named individual participants mentioned in the question
- locations: explicit concrete locations mentioned in the question
- time_range: a pair [start, end] in ISO 8601 format if the question contains a time reference that can be grounded to an absolute date or interval; otherwise null

Rules:
- semantic_query must preserve the original meaning exactly.
- Do not hallucinate missing people, places, or time.
- keywords should be short phrases, not full sentences.
- If no explicit person, location, or time appears, return empty lists or null.

Question: {query}
\end{PromptTemplate}

\Needspace{0.80\textheight}
\begin{PromptTemplate}{Recollection Planning Prompt}
You are deciding whether to continue a graph-based memory recollection step.

Round: {round_name}
Question: {query}
Query analysis: {query_analysis}
Current evidence:
{current_evidence}

Goal:
- Expand only if the current evidence is still insufficient.
- If the current evidence already contains a direct answer or clearly sufficient support, do not expand.
- Treat the current evidence set as the memories that came to mind first.
- Select up to {max_expand_nodes} anchors only from the current evidence set.
- An anchor should be a strong starting point for further recollection: it should either be directly useful for answering the question, or serve as a good local gateway to the missing evidence.
- Do not select all relevant memories; select only the few best recollection starting points.
- Produce exactly one shared evidence_target for this round.
- The evidence_target should describe the missing memory to retrieve next, not the final answer and not a paraphrase of the original question.
- Write target_restatement in the style of a factual memory sentence, similar to a memory node's lossless_restatement, not as a question or retrieval instruction.
- Prefer a concrete missing-evidence description, for example by specifying the missing fact type, participant, location, or time cue when available.
- If time is relevant to the missing evidence, include a normalized ISO 8601 time interval in time_range; otherwise return null.
- Stop when the current evidence is already sufficient, or when further local expansion is unlikely to help.

Example:
Question: When did Caroline and Melanie go to a pride festival together?

Current evidence set:
- memory_a: Caroline attended a pride parade on 2023-08-11 and felt proud and supported.
- memory_b: Melanie said the pride parade looked awesome and asked Caroline if Caroline joined in.
- memory_c: Caroline and Melanie often support each other in LGBTQ-related activities.

A good response:
{
  "continue": true,
  "stop_reason": null,
  "anchors": ["memory_b"],
  "evidence_target": {
    "goal": "find_shared_pride_event_date",
    "target_restatement": "Caroline and Melanie attended a pride-related event together.",
    "persons": ["Caroline", "Melanie"],
    "locations": [],
    "time_range": null
  }
}

Return JSON only:
{
  "continue": true,
  "stop_reason": null,
  "anchors": ["memory_id_1", "memory_id_2"],
  "evidence_target": {
    "goal": "short missing-evidence label",
    "target_restatement": "A concise restatement of the missing evidence to retrieve next.",
    "persons": ["optional explicit people names"],
    "locations": ["optional explicit locations"],
    "time_range": ["ISO_START", "ISO_END"] or null
  }
}
\end{PromptTemplate}

\Needspace{0.50\textheight}
\begin{PromptTemplate}{Answer Generation Prompt}
Answer the user's question based on the provided context.

User Question: {query}

Relevant Context:
{context_str}

Requirements:
1. First, think through the reasoning process.
2. Then provide a very concise answer.
3. Answer must be based only on the provided context.
4. All dates in the response must be formatted as DD Month YYYY when dates are required.
5. Return your response in JSON format.

Output Format:
{
  "reasoning": "Brief explanation of the reasoning process.",
  "answer": "Concise answer."
}

Example:
Question: "When will they meet?"
Context: "Alice suggested meeting Bob at 2025-11-16T14:00:00..."

Output:
{
  "reasoning": "The context explicitly states the meeting time as 2025-11-16T14:00:00.",
  "answer": "16 November 2025 at 2:00 PM"
}

Now answer the question. Return ONLY the JSON, no other text.
\end{PromptTemplate}

\Needspace{0.42\textheight}
\begin{PromptTemplate}[unbreakable]{LLM-as-a-Judge Prompt}
Your task is to label an answer to a question as CORRECT or WRONG.

You will be given:
(1) a question,
(2) a gold answer,
(3) a generated answer.

The point of the question is to ask about something one user should know about the other user based on their prior conversations.

The gold answer will usually be concise and include the referenced topic. The generated answer may be longer. Mark the generated answer as CORRECT if it matches the same core information as the gold answer.

For time-related questions, accept equivalent date or time expressions when they refer to the same time period. Even if the format differs, consider it CORRECT if it refers to the same date or time.

Question: {question}
Gold answer: {gold_answer}
Generated answer: {generated_answer}

Return JSON only:
{
  "label": "CORRECT" or "WRONG"
}
\end{PromptTemplate}

\end{document}